# Les entités spatiales dans la langue:

## étude descriptive, formelle et expérimentale de la catégorisation

Michel Aurnague, Maya Hickmann, Laure Vieu

**Abstract**

While previous linguistic and psycholinguistic research on space has mainly analyzed spatial relations, the studies reported in this paper focus on how language distinguishes among spatial entities. Descriptive and experimental studies first propose a classification of entities, which accounts for both static and dynamic space, has some cross-linguistic validity, and underlies adults' cognitive processing. Formal and computational analyses then introduce theoretical elements aiming at modelling these categories, while fulfilling various properties of formal ontologies (generality, parsimony, coherence…). This formal framework accounts, in particular, for functional dependences among entities underlying some part-whole descriptions. Finally, developmental research shows that language-specific properties have a clear impact on how children talk about space. The results suggest some cross-linguistic variability in children's spatial representations from an early age onwards, bringing into question models in which general cognitive capacities are the only determinants of spatial cognition during the course of development.

**I. Introduction**

Les travaux sur l'expression de l'espace se sont, jusqu'à ce jour, principalement centrés sur les «relations spatiales» exprimées par certains marqueurs (prépositions, postpositions, cas, verbes, etc.) dont le contenu sémantique décrit le lien ou la configuration (statique ou dynamique) s'établissant entre une entité à localiser et une entité de référence (cf. Boons 1987, Herskovits 1986, Svorou 1994, Talmy 2000, Vandeloise 1986, 1988). Les chercheurs ont, comparativement, montré peu d'intérêt pour les éventuelles distinctions entre entités effectuées dans la langue et la cognition sur la base de critères spatiaux. Hormis quelques notions ayant donné lieu à une abondante littérature (ex: massif/comptable, singulier/collectif; cf. Link 1983, Parsons 1970), de nombreuses oppositions, récurrentes dans les travaux linguistiques, sont souvent traitées comme allant de soi (animé/non animé, naturel/artefact, aliénable/inaliénable, mobile/immobile, etc.). Ces dernières, de même que d'autres concepts parfois empruntés à la métaphysique (objet, substance, concret/abstrait, matériel/immatériel, etc.), ne sont pas véritablement discutés sur un plan linguistique et expérimental, ni définis d'un point de vue formel. On est donc en droit de se demander si la langue classifie véritablement les entités spatiales. De plus, et quand bien même cela serait le cas, quels sont les concepts, outils ou processus qui président à une telle catégorisation dans la cognition humaine?

C'est dans le but de structurer le champ d'étude ainsi défini et d'apporter un début de réponse aux questions soulevées qu'a été élaboré le projet «Les entités spatiales et leur catégorisation dans la langue et la cognition». Soutenu par l'ACI Cognitique (COG 135, 2000-2001) et rassemblant une vingtaine de chercheurs —linguistes, psycholinguistes, psychologues du développement, philosophes du langage, logiciens et informaticiens—[1], ce projet a abordé la problématique de la catégorisation des entités spatiales d'un triple point de vue. Des analyses linguistiques détaillées se sont attachées à mettre en évidence et à saisir précisément les distinctions entre entités réalisées par certains marqueurs spatiaux (du français mais aussi du basque, du coréen et du serbo-croate). Diverses expérimentations psycholinguistiques sont venues compléter ces analyses afin de tester leur validité et leur traduction éventuelle en termes de traitement par les locuteurs. Le second volet du projet a comparé la catégorisation issue des observations linguistiques avec les «ontologies» proposées dans d'autres domaines (linguistique formelle, IA, philosophie…) tout en considérant quels outils formels pourraient le mieux rendre compte des distinctions mises au jour. Cette modélisation de l'ontologie spatiale dans la langue a ensuite été utilisée dans le cadre de la formalisation des relations de partie à tout ou méronomies. Un troisième ensemble d'études s'est penché sur la dimension développementale de la catégorisation, en examinant la manière dont les concepts spatiaux concernés se mettent en place chez les nourrissons et les jeunes enfants (avec ou sans troubles de la motricité). L'une des questions centrales de ces travaux expérimentaux porte sur les rôles respectifs de la langue et du reste de la cognition dans l'émergence et le fonctionnement des catégories étudiées. Ces trois facettes complémentaires des recherches sur la catégorisation des entités spatiales sont présentées, tour à tour, dans la suite.

---

[1] Equipes participantes: ERSS (UMR 5610, CNRS et Univ. Tlse 2), IRIT (UMR 5505, CNRS et Univ. Tlse 3), Laboratoire Cognition et Développement (UMR 8605, CNRS et Univ. Paris 5), Laboratoire Jacques Lordat (EA 1941 et INSERM U455, Univ. Tlse 2).

**II. La catégorisation des entités à la lumière des données linguistiques**

Les analyses sémantiques des relations locatives s'attachent généralement à déterminer les contraintes que doit satisfaire une configuration spatiale constituée par une entité à localiser («entité-cible») et une entité de référence («entité-site») afin d'être décrite par un marqueur donné. Ainsi que nous l'illustrons ci-dessous, l'explicitation des configurations adéquates révèle, bien souvent, d'importantes propriétés des entités spatiales en présence.

L'étude des emplois statiques de la préposition *à* du français (c'est-à-dire des emplois qui ne font pas appel au déplacement; cf. Vandeloise 1988) a montré que les sites sélectionnés par ce marqueur (dans ses usages «localisateurs») doivent remplir une «fonction de localisation» c'est-à-dire occuper une position qui soit connue dans la connaissance partagée des locuteurs. Un examen plus poussé (Aurnague 2004, Borillo 2001) indique que les sites concernés (qualifiés de «lieux») se caractérisent par leur stabilité ou fixité dans un cadre de référence approprié (on notera cette propriété +fix) ainsi que par leur capacité à déterminer des «portions d'espace» dans lesquelles des entités-cibles peuvent être localisées (+esp). Ces propriétés sont vérifiées par les lieux géographiques (ex: *Max est à Toulouse, Max est au jardin public*) qui ont une position fixe au sein du cadre de référence terrestre et ne sont pas strictement «matériels» puisqu'ils définissent des portions d'espace contiguës à leur surface au sol (un oiseau volant à travers un pré pourra être décrit comme se trouvant *dans ce pré*). Mais elles sont également valides pour les Noms de Localisation Interne (NLI; ex: *Max est à l'**extrémité** de la table, Le lustre est au-**dessus** de l'armoire*) dans la mesure où ceux-ci dénotent des parties stables à l'intérieur d'une «entité-tout» et identifient, bien souvent, des portions d'espace associées à ces parties matérielles (dans les exemples précédents comme dans *Il y a des fleurs au pied du lampadaire* ou *La mouche est sur le coin du tapis*, Max, le lustre, les fleurs ou la mouche peuvent être localisés à une certaine distance des sites mentionnés). Outre qu'ils répondent aux conditions caractérisant la notion de lieu —fixité de l'élément désigné (+fix) et présence d'une portion d'espace (+esp)—, noms propres d'entités géographiques et NLI présupposent (pour les premiers) ou apportent explicitement (pour les seconds) des informations sur la localisation de l'entité dénotée (+spc: on a donc affaire à des «lieux spécifiés»), confirmant ainsi les observations initiales de C. Vandeloise. Mais les contraintes sous-tendant la notion de lieu et celles liées à la spécification doivent être distinguées. En effet, certaines expressions peuvent introduire des lieux sans en préciser ou présupposer la position (emploi de l'indéfini), ce qui conduit au rejet des constructions considérées (ex: *\*Max est à un jardin public*). Inversement, il suffit que l'une des deux propriétés caractéristiques des lieux (et a fortiori les deux) soit violée pour que l'usage de la préposition *à* devienne inacceptable et ceci malgré le recours à un article défini supposé garantir l'aspect spécifié (ex: *\*Max est au rocher* (+fix,-esp), *\*La mouche est au verre* (-fix,+esp), *\*La mouche est au couteau* (-fix,-esp)). Les entités entrant dans ce cas de figure (non fixité ou absence de portions d'espace) seront qualifiées d'«objets».

Ce bref examen des emplois localisateurs de *à* nous a permis d'évoquer les notions de «lieu», de «portion d'espace» et d'«objet». Les résultats d'autres travaux sémantiques pourraient compléter ce panorama. C'est le cas de la préposition *dans* (Vandeloise 1986, Vieu 1991) dont le fonctionnement fait appel aux classes mentionnées ci-dessus ainsi qu'aux notions de «substance», de «portion de matière» ou de «collection». De même, l'étude des génitifs locatif et possessif du basque (lorsqu'ils réfèrent à des relations entre parties fonctionnelles et tout) montre que la distribution de ces marqueurs est indirectement conditionnée par les concepts de lieu, d'objet et de bâtiment ou «entité mixte» (Aurnague 2004). S'ils vérifient bien les propriétés associées aux lieux (+fix,+esp), les bâtiments présentent, en effet, certaines caractéristiques propres aux objets (ex: décomposition en parties bien circonscrites): les descriptions spatiales pourront, dès lors, se focaliser sur l'un ou l'autre de ces aspects (nom propre: lieu géographique, nom commun: lieu géographique ou objet). Ainsi que le suggère cette dernière remarque, la classification des entités sous-tendant les descriptions spatiales ne doit pas être conçue comme quelque chose de rigide/immuable (et extérieur au système linguistique) mais comme un «point de vue» adopté par la langue au moment d'identifier une entité (point de vue qui est, bien sûr, conditionné par les propriétés du référent considéré).

Les catégories «ontologiques» ainsi dégagées par l'analyse sémantique conduisent à se poser deux types de questions. Si de telles distinctions paraissent bel et bien intervenir dans le fonctionnement des marqueurs linguistiques, ont-elles des implications plus ou moins directes en termes de traitement par les sujets ou locuteurs? Par ailleurs, les classes d'entités définies pour l'espace statique sont-elles compatibles avec les notions mises en jeu par l'expression du déplacement?

La première des deux interrogations a été appréhendée à travers une série d'expérimentations psycholinguistiques visant à mieux saisir le traitement, par des sujets normaux, de certains «noms de parties». Il s'agit, d'une part, des Noms de Localisation Interne (NLI; ex: *avant, dessus, coin, fond, extrémité*) dont on a vu qu'ils désignent des «lieux spécifiés» (+fix, +esp, +spc) et, d'autre part, des noms de composants dénotant des parties fonctionnelles d'entités (ex: *anse, couvercle, tiroir, roue, voyant*). Bien qu'ils délimitent des parties stables d'un tout (+fix), les noms de composants se distinguent des NLI par le fait qu'ils n'identifient pas de portions d'espace associées aux zones matérielles isolées (-esp: ils réfèrent à des objets plutôt qu'à des lieux) et indiquent la fonction de la partie concernée plutôt que sa localisation (-spc). Les expérimentations réalisées ont toutes consisté en des tâches de pointage: la représentation d'une entité spatiale, accompagnée d'un NLI ou d'un nom de composant, est affichée sur un écran d'ordinateur, le sujet/locuteur devant désigner (avec son index) la partie de l'entité qui lui paraît correspondre au substantif (les coordonnées de la zone pointée et le temps mis pour l'exécution sont enregistrés). Les tests les plus récents ont révélé que le recours à des entités «complexes» en termes de structuration interne (bâtiments divers, chaîne stéréo, tableau de bord (de voiture), etc.) implique des traitements significativement plus longs pour les noms de composants que pour les NLI. Cet allongement des temps de réponse dans le cas des composants n'apparaît pas pour les entités plus simples (ex: théière, lampe, valise). Il est donc lié à la complexité des entités. Ces données confirment, d'une certaine manière, l'hypothèse initiale selon laquelle les noms de composants auraient un sémantisme fonctionnel (-spc) contrairement aux NLI qui seraient éminemment localisateurs (+spc) et se verraient, dès lors, moins affectés par les variations de structure interne. Au cours de tests antérieurs portant sur les NLI, l'examen des nuages de points enregistrés pour chaque stimulus a mis en évidence le fait que les sujets désignent assez régulièrement des zones (non matérielles) proches des parties découpées par les NLI sur les entités, ce qu'on ne peut manquer de mettre en relation avec la capacité de ces marqueurs à identifier des portions d'espace (+esp). L'analyse des pointages correspondant aux noms de composants reste à faire mais il est probable qu'elle révèle une plus grande coïncidence avec les parties matérielles des entités (-esp: les noms de composants désignent des objets). Au-delà de l'opposition lieu (spécifié)/objet, ces expérimentations ont aussi permis d'éclairer le fonctionnement des NLI orientationnels (ex: *avant*, *dessus*, *gauche*). Il a ainsi été montré que le traitement de l'orientation frontale (*avant/devant*) fait appel à deux facteurs basés sur les propriétés intrinsèques des entités —fonction liée au mouvement ou indépendante de celui-ci— ainsi qu'à deux facteurs contextuels —saillance géométrique et aérodynamie—, seuls les premiers étant susceptibles d'intervenir dans la classification des entités (Aurnague, Champagne, Vieu, *et al.* à paraître).

Pour conclure cette section, et en réponse au second point soulevé plus haut, nous dirons que l'étude des prépositions *par* et *à travers* du français semble indiquer que si les descriptions dynamiques ne remettent pas en cause la catégorisation proposée pour l'espace statique, elles introduisent des distinctions permettant de la compléter. C'est le cas des «conduits» et des «chemins» dont on peut montrer qu'ils constituent, respectivement, des sous-classes des objets et des lieux (Aurnague et Stosic 2002).

**III. Représenter les catégories et leurs propriétés formelles**

Si l'on adopte un point de vue référentiel sur le langage, il convient d'expliquer comment les catégories mises au jour en linguistique descriptive sont prises en compte dans un cadre général de représentation de la sémantique des langues. La sémantique formelle a traditionnellement pour objet d'établir le lien entre forme et sens, plus particulièrement, entre structures syntaxiques et structures sémantiques d'une langue, et suppose pour cela un cadre logique de représentation du monde[2] auquel la langue réfère. Un tel cadre ne rend compte, en général, que des aspects du monde exprimés à travers la syntaxe des langues les plus étudiées —les langues indo-européennes—, sans chercher à constituer une ontologie complète. Ces aspects ne se limitent cependant pas aux catégories logiques fondamentales telles que prédicat, référent et quantificateur: l'étude de phénomènes linguistiques comme la détermination et l'expression du temps a mené à l'enrichissement du cadre logique à l'aide de théories caractérisant certaines catégories d'entités. En ce qui concerne les catégories impliquant les entités spatiales, les travaux sur les termes de masse ont produit des théories des «quantités de matière» et des «substances» (Parsons 1970) et les travaux sur les termes comptables et le pluriel ont donné lieu à des théories des entités plurielles et des collections (Link 1983). Au-delà des structures syntaxiques, la prise en compte du lexique, notamment spatial, pour «remplir» les formules sémantiques suppose que l'on étende encore ce cadre significativement. Malgré certains travaux poussés sur la structure du lexique (ex: Pustejovsky 1995), aucune théorie existante et caractérisant en profondeur (c'est-à-dire axiomatiquement, pour supporter des raisonnements) les diverses catégories d'entités ne cherche à la fois à s'insérer dans le cadre de la sémantique formelle et à couvrir un large champ lexical.

---

[2] Ce terme est volontairement imprécis. Ce monde peut être la «réalité» aussi bien qu'une représentation cognitive de la réalité.

Nous nous sommes donc penchés, dans ce projet, sur ce que propose la littérature en ontologie formelle qui vise à élaborer des théories plus globales de représentation du monde, même si il n'y a certainement pas, dans cette branche de la métaphysique (étude de la nature et de la structure de la réalité), de consensus sur une catégorisation donnée. Par exemple, Casati et Varzi (1999) axiomatisent une relation de localisation à partir de laquelle ils définissent ce que sont les entités concrètes, et parmi celles-ci, distinguent les entités purement spatiales (les «régions») des autres. D'autres travaux, basés sur les relations fondamentales que sont la relation méréologique de partie à tout (Simons 1987) et les relations de dépendance (Fine 1995), permettent de caractériser les liens entre différents «substrats» (espace, temps, matière). Cependant, du fait que l'objet d'étude de la métaphysique est «la» réalité, dont cette discipline postule l'existence, l'ensemble de ces travaux d'ontologie formelle reste largement indépendant de la langue et de la cognition.

Aucune des propositions théoriques classiques ne permettant donc de représenter formellement l'ensemble des catégories d'entités spatiales identifiées en langue, nous avons cherché à rassembler les propositions les plus adéquates en un seul cadre théorique cohérent. Les principaux écueils à affronter pour produire une théorie qui rende compte des catégories linguistiques tout en satisfaisant les critères de généralité, de parcimonie, de cohérence et de bien-fondé propres à l'ontologie formelle sont le contrôle de la multiplication d'entités co-localisées et la prise en compte de la notion de contexte, essentielle pour certaines catégories linguistiques. Pour éviter ces écueils, le cadre général de formalisation élaboré au cours de nos travaux (Aurnague, Vieu et Borillo 1997, Gambarotto et Muller 2003) est stratifié. Il distingue le substrat spatio-temporel, ne correspondant pas à des entités directement descriptibles linguistiquement, mais à leur composante purement spatio-temporelle (Muller à paraître), du niveau des entités langagières, dans lequel on différencie quatre catégories fondamentales d'entités: les entités matérielles, les éventualités, les substances, et les entités immatérielles ou «portions d'espace». Ces quatre catégories ne sont caractérisées que partiellement par les relations et propriétés méréotopologiques —par exemple l'inclusion, le contact ou la propriété d'être connexe, c'est-à-dire d'un seul tenant— axiomatisées au niveau du substrat spatio-temporel. Diverses sortes de relations de dépendance (constitution, participation, quantité…) sont introduites et formalisées au niveau des entités langagières afin de définir pleinement les quatre catégories. Cette catégorisation est complétée par les notions d'entité singulière, d'entité plurielle et de collection, applicables indistinctement aux quatre catégories. La catégorie des entités matérielles comprend la sous-catégorie linguistiquement pertinente des objets (artéfactuels, biologiques ou non), les «portions de matière» (désignés par des termes de masse), les lieux et les «entités mixtes» (bâtiments, cf. section II). Ces deux dernières sous-catégories sont plus complexes à caractériser car la notion de lieu n'est pas réellement ontologique, puisqu'elle est fondée sur la donnée d'un cadre de référence, c'est-à-dire d'un contexte ou d'un point de vue. La solution adoptée consiste en la définition de cadres de référence, ensembles d'entités —les lieux— munies de relations spatiales qui caractérisent leur fixité relative pendant une période donnée et telles que chacune détermine une portion d'espace associée (cf. section II). Ces travaux trouvent actuellement un prolongement et une généralisation grâce à l'introduction, en collaboration avec N. Guarino et son équipe, de la notion de concept relatif à un contexte pour le développement d'ontologies cognitives et sociales (Masolo *et al.* 2004).

Ce cadre théorique de représentation des catégories d'entités spatiales a été ensuite mis à l'épreuve pour tester son utilité dans la représentation des relations de partie à tout, sous-jacentes à un grand nombre d'expressions spatiales, que ce soit à travers l'usage de Noms de Localisation Interne ou bien de noms de composants (cf. section II). Une étude systématique visant à recenser les expressions du français et du basque permettant de décrire une relation de partie à tout a été tout d'abord réalisée (Aurnague 2004). Ces relations, telles qu'elles s'expriment en langue, sont beaucoup plus riches sémantiquement que la simple relation méréologique de base, qui ne caractérise en fait guère plus que l'inclusion spatio-temporelle. En particulier, de nombreux auteurs, en linguistique et psycholinguistique, ont constaté des cas d'intransitivité (Winston, Chaffin et Herrmann 1987, Iris, Litowitz et Evens 1988), ce qui les a amenés à distinguer différentes relations de partie à tout, sans toutefois les formaliser. L'étude linguistique nous a permis de reconsidérer les diverses classifications proposées et a fait clairement apparaître que la catégorisation des entités joue un rôle important dans la distinction entre relations.

Le cadre formel ontologique que nous avons développé permet effectivement de rendre compte de la contrainte d'homogénéité de la catégorie entre partie et tout, et également de plusieurs relations. On définit en effet aisément cinq relations: «membre-collection» (*une brebis du troupeau*), «sous-collection-collection» (*le couple Dupont fait partie des heureux gagnants*), «portion-tout» (*une part de gâteau*), «substance-tout» (*la farine du gâteau*) et «morceau-tout» (*le haut de la pièce, un morceau de la tasse*). La relation principalement signalée dans la littérature, mais rarement formalisée, «composant-tout intégré» (*la main de jean, la poignée de la maison...*) manque toutefois. La notion de fonctionnalité, caractéristique principale de cette relation, est complexe. Les propositions non téléologiques de la littérature faisant souvent appel à la notion même de composant (Cummins 1975), nous avons cherché à sortir de ce cercle vicieux en montrant combien la fonctionnalité est liée à la notion de dépendance. Nous avons finalement proposé une axiomatisation des dépendances fonctionnelles générique et individuelle et défini quatre sortes de relation «composant-tout intégré», en fonction du sens de la dépendance entre la partie et le tout et en fonction de la nature directe ou indirecte de cette dépendance, ce qui nous a permis d'expliquer les comportements inférentiels apparemment irréguliers de cette relation (Vieu et Aurnague à paraître).

**IV. Les relations spatiales et la catégorisation chez l'enfant**
Une question centrale soulevée par les recherches descriptives et formelles est celle du développement. Quand et comment les catégories d'entités spatiales mises en évidence par ces approches apparaissent-elles chez l'enfant? Comment évoluent-elles au cours du développement de la cognition spatiale? Quels facteurs (perceptifs, cognitifs, linguistiques) peuvent rendre compte de cette évolution? En fait, la représentation de l'espace chez l'enfant fait actuellement l'objet de vives controverses opposant différentes conceptions «universalistes» et «relativistes» de la cognition spatiale. Un premier ensemble de modèles universalistes, inspiré de la théorie piagetienne, postule que l'acquisition du langage spatial serait exclusivement déterminée par le développement sensori-moteur et cognitif, selon des étapes universelles et indépendantes du langage. Par ailleurs, de nombreuses recherches (Lécuyer, Streri et Pêcheux 1996) montrent que le nourrisson fait preuve d'une connaissance complexe et précoce du monde physique qui l'entoure, amenant certains modèles à postuler l'existence de capacités innées et modulaires, alors que d'autres accordent un rôle prépondérant à l'activité perceptive du bébé. Enfin, un nombre croissant d'études (Bowerman 1996, à paraître, Bowerman et Choi 2003, Hickmann 2002, 2003, Hickmann, Hendriks et Roland, 1998, Slobin 1996, à paraître) montrent que la langue a une incidence sur le développement, notamment sur la représentation de l'espace. De tels résultats vont à l'encontre des hypothèses universalistes, amenant certains à postuler que l'enfant est sensible aux propriétés de la langue dès le plus jeune âge et que celles-ci infléchissent le développement de la cognition spatiale.

Dans le contexte de ce débat, nos recherches (Hickmann à paraître, Hickmann et Hendriks à paraître) ont examiné l'expression des relations spatiales, d'abord en français (adultes et enfants de 3 à 6 ans), langue encore quasiment inexplorée dans ce domaine du développement, puis auprès d'adultes anglophones. Les sujets ont participé à deux tâches, l'une statique, l'autre dynamique. Dans la première, ils devaient localiser une cible par rapport à un site à partir d'une soixantaine d'images montrant différents types de configurations. Certaines visaient l'emploi de prépositions particulières, notamment *sur/on* (une tasse sur une table), *dans/in* (une pomme dans un bol)*, sous/under* (un ballon sous une chaise), *au-dessus/above* (un nuage au-dessus d'une maison). D'autres items

«exploratoires» étaient plus complexes, par exemple parce que le support nécessaire au contact n'était pas horizontal et/ou impliquait une certaine manière d'attachement de la cible par rapport au site (une affiche sur un mur, une veste accrochée à un portemanteau). Dans la seconde tâche, les sujets devaient décrire une vingtaine d'actions («mettre», puis «enlever») consistant à déplacer des objets cibles dans différentes localisations qui impliquaient un contenant (des jouets dans un sac), une partie du corps (une veste sur une poupée) ou d'autres entités (un couvercle sur une casserole), parmi lesquelles certaines comportaient une relation étroite avec la cible (un légo dans un autre).

Globalement, les adultes codent l'information nettement plus souvent dans les particules et les prépositions en anglais, mais dans les verbes en français. En français, notamment dans la tâche dynamique et avec les items statiques exploratoires, l'emploi de prépositions neutres (*à, de*) et l'absence de préposition sont d'autant plus fréquents que les verbes contiennent des informations spécifiques (par exemple, *mettre/être sur le crochet* vs. *accrocher/accroché au crochet; mettre un légo dans un autre* vs. *emboîter des légos*). Par ailleurs, les informations qui sont véhiculées par les verbes sont différentes dans les deux langues. En français, les sujets se focalisent surtout sur la manière d'attachement de la cible au site (*accrocher/décrocher*, *emboîter/désemboîter, encastrer, enfiler,* etc.). En revanche, si les sujets anglophones codent parfois l'attachement, ils le font moins souvent et de façon très générale (*to fix, to attach, to connect*), se focalisant plutôt sur d'autres types d'informations, telle la posture (*to sit, to hang*) ou la manière d'effectuer les changements de localisation (*to press, to pull*). Enfin, dès 3 ans, les réponses des enfants français sont clairement conformes à celles des adultes français. Ainsi, à tous les âges, il existe une relation étroite entre verbes et prépositions, et les enfants utilisent souvent le verbe pour coder la manière d'attachement. Néanmoins, les réponses évoluent également avec l'âge de deux manières. Tout d'abord, jusqu'à 5 ans, on observe deux types de sur-généralisations de la préposition *sur*: des emplois inappropriés sur le plan sémantique, notamment avec les items visant à induire la préposition *au-dessus* (*Le nuage est sur la maison* pour décrire une relation n'impliquant pas de contact nuage-maison); des emplois de *sur* avec les items statiques exploratoires, appropriés sur le plan sémantique, mais non attestés chez les adultes français, qui utilisent le plus souvent des prépositions neutres ou des verbes sans prépositions dans ces cas (*être sur le crochet* vs. *être suspendu au crochet* ou *être accroché [au crochet]*). Par ailleurs, entre 3 et 6 ans, l'emploi de verbes spécifiques augmente avec ces items, ainsi que dans la tâche dynamique (*mettre sur le crocher* vs. *suspendre au crochet* ou *accrocher [au crochet]*)

L'analyse détaillée des réponses montre que celles-ci reflètent une «ontologie» sous-jacente, qui est partiellement semblable et partiellement variable d'une langue à l'autre. Ainsi, l'emploi des verbes spécifiques est fortement lié aux propriétés des entités (cf. sections II et III). Par exemple, ces verbes sont les plus fréquents dans les deux langues lorsque les entités sont étroitement liées entre elles de par leur forme et/ou leur fonction (*emboîter des légos, accrocher un vêtement, enfiler des perles, recapuchonner un stylo, panser une plaie,* etc.). Néanmoins, en français, c'est souvent le seul verbe qui code l'information (*emboîter/désemboîter des légos*), alors qu'en anglais les particules et les prépositions fournissent le plus souvent des informations partielles (*to plug legos into each other, to pop legos apart*). De plus, les locuteurs français codent les propriétés des entités par le biais du verbe (*chausser/déchausser la poupée, coiffer la poupée d'un chapeau, panser la main, recapuchonner le stylo*) dans de nombreux cas où les locuteurs anglophones ne représentent pas ces propriétés et utilisent les particules et les prépositions pour coder la localisation (*to put a shoe/hat onto the doll, to put a band aid on the hand, to put the top onto the pen*). Enfin, des réponses particulières sont produites en français lorsque les entités présentent certains types de relations partie-tout (la poignée d'une porte de placard) ou des entités immatérielles (une fissure, un trou) (cf. sections II et III). Dans ces cas, contrairement aux locuteurs anglophones qui localisent les entités cibles (*a handle on a door, a crack in a cup, a hole in the towel*), les sujets francophones de tous les âges tendent à les identifier en exprimant une relation de type partie-tout (*une poignée de porte*) ou en attribuant des propriétés au site (*une tasse fêlée, un torchon troué*). Des analyses dans d'autres langues sont nécessaires afin de déterminer dans quelle mesure de telles réponses sont spécifiques au français.

D'un point de vue développemental, ces résultats montrent une complémentarité entre deux types de déterminants: ceux qui sont liés au développement cognitif général (les processus de sur-généralisation et d'expansion lexicale); ceux qui dépendent des propriétés typologiques de la langue (le rôle plus ou moins central du verbe ou d'autres éléments) entraînant un recours plus ou moins important à la lexicalisation ou à la grammaticalisation (Talmy 2000) et, de ce fait, une focalisation sur différents aspects de l'espace. L'impact conjoint de ces deux types de déterminants entraîne chez l'enfant des réponses qui sont conformes au système adulte et cette conformité est à la fois précoce (dès 3 ans) et croissante (de 3 à 6 ans et vraisemblablement au-delà). Ces résultats remettent en question l'idée que le développement de capacités cognitives générales serait le seul déterminant de la cognition spatiale. Les modèles les plus récents proposent que la langue sert de filtre qui canalise l'information, incitant les locuteurs à prêter plus ou moins d'attention à différents aspects de la réalité. Plusieurs types d'études en cours poursuivent ces recherches dans différentes directions. Certaines (Lécuyer et Rivière à paraître) montrent que les jeunes enfants atteints de troubles de la motricité ne présentent aucun retard dans l'utilisation des prépositions spatiales, démontrant ainsi que la cognition spatiale ne dépend pas du développement moteur, mais plutôt de la perception. Des études longitudinales comparatives chez le jeune enfant testent également l'hypothèse que la langue a une incidence sur les processus de mise en place des représentations spatiales dès l'émergence du langage. Ces études impliquent l'analyse de productions spontanées recueillies auprès des mêmes enfants de façon régulière et à des intervalles rapprochés au cours de plusieurs années (tous les mois entre 18 mois et 4 ou 5 ans), et ceci dans plusieurs langues (romanes et germaniques). Enfin, des recherches effectuées auprès de nourrissons (dès 9 mois) examinent la compréhension des marqueurs spatiaux en relation avec la catégorisation des entités spatiales avant même l'émergence du langage.

**V. Conclusion**

En guise de bilan, nous dirons que les travaux descriptifs effectués dans le cadre de ce projet indiquent que de nombreux marqueurs de l'espace font appel à une catégorisation sous-jacente des entités spatiales, certains des concepts ou oppositions mis en évidence (ex: objet, lieu) paraissant avoir une validité interlinguistique (outre les données du français et du basque déjà évoquées, on peut citer les recherches sur le coréen (Choi-Jonin et Sarda à paraître) et le serbo-croate (Stosic 2002)). La classification ainsi opérée par la langue n'a cependant pas un caractère rigide/immuable et permet d'appréhender une même entité selon des points de vue divers.

Une première modélisation de ces distinctions linguistiques a été réalisée en s'inspirant des outils élaborés en ontologie formelle et des propriétés qui leur sont habituellement associées (généralité, parcimonie, cohérence…). Les représentations proposées distinguent les entités directement identifiées par la langue de leur substrat ou composante spatio-temporel et s'attachent à prendre en considération le rôle du contexte ainsi que les liens entre éléments co-localisés. Cet appareillage formel s'avère extrêmement utile pour la formalisation des relations de partie à tout.

Enfin, l'étude développementale des descriptions spatiales montre que celles-ci sont à la fois conditionnées par des mécanismes perceptifs et cognitifs généraux et par les propriétés typologiques des langues. Ces dernières jouent le rôle de filtres orientant le locuteur vers des notions et catégories particulières et le conduisant à privilégier tel ou tel aspect de son environnement. Ceci suppose une certaine variabilité translinguistique des catégories d'entités, du moins dans le fonctionnement habituel des locuteurs dès le plus jeune âge. Ces résultats remettent en question les présupposés de certains modèles, notamment l'idée que le développement de capacités générales serait le seul déterminant de la cognition spatiale.

**Références bibliographiques**

Gentner et Susan Goldin-Meadow (éds.), *Language in Mind: Advances in the study of language and thought*. Cambridge (MA), MIT Press: 387-427.

Casati, Roberto et Achille Varzi. 1999. *Parts and Places - The Structures of Spatial Representation*. Cambridge (MA), MIT Press.

Choi-Jonin, Injoo et Laure Sarda. À paraître. «The expression of Motion, Path and Deixis components in orientation motion verbs: a cross-linguist account based on French and Korean», in Michel Aurnague, Maya Hickmann et Laure Vieu (éds), *The categorization of spatial entities in language and cognition*.

Cummins, Robert. 1975. «Functional Analysis», *Journal of Philosophy*, 72: 741-765.

Fine, Kit. 1995. «Ontological Dependence», *Proceedings of the Aristotelian Society*, 95: 269-90.

Gambarotto, Pierre et Philippe Muller. 2003. «Ontological problems for the semantics of spatial expressions in natural language», in Emile van der Zee et Jon Slack (éds.), *Representing direction in language and space*. Oxford: Oxford University Press: 144-165.

Herskovits, Annette. 1986. *Language and spatial cognition: an interdisciplinary study of the prepositions in English*. Cambridge, Cambridge University Press.

Hickmann, Maya. 2002. «Espace, langage et catégorisation : le problème de la variabilité inter-langues», in Jacques Lautrey, Bernard Mazoyer et Paul van